%% file: handbook03.tex
\documentclass{svmult}

\usepackage{makeidx}     
\usepackage{graphicx}    
\usepackage{multicol}    

\makeindex             

\begin{document}

\input{mydefs}

 \newenvironment{my-quote}{                 
      \begin{quote} 
	\parskip -0.1cm \topsep -0.1cm}
        {\end{quote} \normalsize} 
\title{Knowledge Patterns}
\titlerunning{Knowledge Patterns}
\author{Peter Clark\inst{1} \and John Thompson\inst{1} \and Bruce Porter\inst{2}}
\authorrunning{Peter Clark, John Thompson, and Bruce Porter}
\institute{Knowledge Systems, \\
Boeing Mathematics and Computing Technology  \\
MS 7L66, PO Box 3707, Seattle, WA 98124 \\
\texttt{\{peter.e.clark,john.a.thompson\}@boeing.com} 
\and
Computer Science Dept.\\
University of Texas \\
Austin, TX 78712 \\
\texttt{porter@cs.utexas.edu}}

\maketitle

\begin{abstract}
This Chapter describes a new technique, called ``knowledge patterns'', for helping
construct axiom-rich, formal ontologies, based on identifying and explicitly
representing recurring patterns of knowledge (theory schemata) in the ontology,
and then stating how those patterns map onto domain-specific concepts in the
ontology. From a modeling perspective, knowledge patterns provide an important
insight into the structure of a formal ontology: rather than viewing a formal
ontology simply as a list of terms and axioms, knowledge patterns views it
as a collection of abstract, modular theories (the ``knowledge patterns'') plus
a collection of modeling decisions stating how different aspects of the world
can be modeled using those theories. Knowledge patterns make both those abstract
theories and their mappings to the domain of interest explicit, thus making
modeling decisions clear, and avoiding some of the ontological confusion that
can otherwise arise. In addition, from a computational perspective, knowledge
patterns provide a simple and computationally efficient mechanism for facilitating
knowledge reuse. We describe the technique and an application built using them,
and then critique its strengths and weaknesses. We conclude that this technique
enables us to better explicate both the structure and modeling decisions made
when constructing a formal axiom-rich ontology.
\end{abstract}

\section{Introduction}
At its heart, ontological engineering is a modeling endeavor. In a formal
ontology, in particular, the knowledge engineer attempts to identify concepts 
and axioms which reflect (to a certain approximation) the real-world phenomena which
he/she is interested in. A common observation is that, when doing
this, one often finds oneself repeating structurally similar patterns of
axioms. For example, when formalizing an ontology about a
space science experiment (called KB-PHaSE \cite{phase}),
we found that axioms about connectivity in electrical circuits, and 
about connectivity in optical systems, had substantial structure in common.
To make this shared structure explicit, and hence reusable, we have
developed a knowledge engineering technique based on the explicit 
representation of these {\it knowledge patterns}, i.e., general 
templates denoting recurring theory schemata, and their transformation
(through symbol renaming) to create specific theories, which we
present in this Chapter.

From a knowledge engineering point of view, knowledge patterns
provide considerable flexibility, as they can be transformed in multiple
ways, and can be used in whole or in part. We describe how this overcomes
some of the limitations of trying to use inheritance to achieve the
same effect. From a philosophical point of view, knowledge patterns 
are also significant as they provide structure to the knowledge in 
an ontology, explicitly modularizing and separating the abstract 
theories (the knowledge patterns) from the phenomena in the world which
those theories are deemed to reflect. For example, rather than encoding a 
theory about {\it electrical circuits}, we encode a knowledge pattern about 
{\it directed graphs}, and then state how an electrical circuit can be 
{\it modeled as} as a directed graph. In this way, knowledge patterns make 
explicit (and reusable) the ``computational clockwork'' of our axioms, 
and the modeling decisions made to apply that clockwork to the task at hand. 
As a result, a formal ontology can be viewed as a collection of theories 
mapped onto the domain of interest (perhaps in multiple ways), rather than 
simply as a ``sea of axioms''.


%


Consider, for example, constructing a formal ontology about banking.
We might include axioms such as: if an amount X is deposited into a bank
account, then the the amount in that account is increased by X. We 
could write many such axioms, and as a result have a useful theory
about banking. However, what is not represented here is a fundamental --
and perhaps subconscious -- insight by the knowledge engineer, namely
that {\it a bank account can be modeled as a kind of container}, and 
thus that a theory of containers can be applied, in this case, to
bank accounts. The axiom above, for example, asserts a container-like
behavior on bank accounts, but nowhere is the abstract container theory
itself stated, nor the mapping from it to bank accounts made explicit.
Without this insight, the knowledge engineer will 
find him/herself writing the same pattern of axioms many times for
different container-like phenomena. Our goal with knowledge patterns is 
to avoid this by making such abstract theories explicit, distinct from 
their application to a particular domain, and hence reusable. We
aim to separate the ``computational clockwork'' of an axiom set from
the real-world phenomena which (according to the knowledge engineer)
seems to behave in a similar way to that axiom set. 

As another example, consider the various formal ontologies of time, with 
axioms about time points, time intervals, etc. In fact, large parts of these
theories are not specifically about time; rather, they can be viewed
as (in part) as theories about {\it lines}, along with the implicit insight
that ``time can be modeled as a line''. Again, our goal with knowledge
patterns is to make explicit the underlying model (here, of lines), and
its application to some phenomenon (here, time). 

It might seem that this type of reuse could also be achieved using normal
inheritance mechanisms (e.g., asserting ``a bank account isa container'',
or ``time isa line''). However, this works poorly in two situations:
when the abstract theory applies to a specific theory in
more than one way, and when only a selected portion of the abstract
theory is applicable. In the next Section, we discuss in detail an
example to illustrate these problems, and subsequently describe
the knowledge pattern approach, and how it overcomes these
limitations. We conclude that this technique enables us to
better modularize axiom-rich ontologies and reuse their general theories.

\section{The Limitations of Inheritance \label{limits-of-inheritance}}

Consider constructing an ontology about computers, including formal axioms
to define the meaning of the terms and relations used in that ontology.
We might include relations in the ontology such as 
\m{ram\_size} (the amount of RAM a computer has), 
\m{expansion\_slots} (the number of expansion slots a computer has), 
\m{free\_slots} (the number of free slots a computer has), etc.,
and formalize the meaning of these terms 
using axioms such as the following, here expressed in 
Prolog\footnote{Variables
start with upper-case letters and are universally quantified; `\m{:-}'
denotes reverse implication ($\leftarrow$); `\m{,}' denotes
conjunction; and \m{is} denotes arithmetic computation.}:

\begin{my-quote}
{\it ~~~~\% ``Available RAM is the total RAM minus the occupied RAM.''} 
\begin{verbatim}
available_ram(Computer,A) :- 
    isa(Computer,computer),
    ram_size(Computer,S), 
    occupied_ram(Computer,R), 
    A is S - R.

\end{verbatim}
{\it ~~~~\% ``The number of free expansion-slots is the total number} \\
{\it .~~~\% ~~of slots minus the number filled.''} 
\begin{verbatim}
free_slots(Computer,N) :- 
    isa(Computer,computer),
    expansion_slots(Computer,X), 
    occupied_slots(Computer,O),
    N is X - O.
\end{verbatim} \end{my-quote}
\noindent
The two axioms above are syntactically different, yet they both
instantiate the same general axiom, which we could explicate as:
\begin{my-quote} \begin{verbatim}
FREE_SPACE(X,S) :-
    isa(X,CLASS),
    CAPACITY(X,C),
    OCCUPIED_SPACE(X,O),
    S is C - O.
\end{verbatim} \end{my-quote}
As part of a general {\it container} theory, this axiom relates a
container's free space, capacity, and occupied space.

The axioms for \m{available\_ram} and \m{free\_slots} are
instantiations of this axiom just when a computer is modeled as a
container of data and expansion cards, respectively.  However, unless
this general theory of {\it containers} is represented explicitly, its
application to the domain of computers is only implicit.
Clearly, we would prefer to explicitly represent the theory,
then to reuse its axioms as needed.

This is typically done with inheritance.  The knowledge engineer
encodes an explicit theory of {\it containers} at a high-level node in
a taxonomy, then its axioms are automatically added to more specific
theories at nodes lower in the taxonomy. 
One axiom in our {\it container} theory might be:
\begin{my-quote} 
\begin{verbatim}
free_space(Container,F) :- 
    isa(Container,container),
    capacity(Container,C), 
    occupied_space(Container,O), 
    F is C - O.
\end{verbatim} \end{my-quote}

To use inheritance to import this axiom into our {\it computer}
theory, we assert that computers are containers and that \m{ram\_size}
is a special case (a `subslot,' in the terminology of frame systems)
of the \m{capacity} relation:


\begin{my-quote}
{\it \% ``Computers are containers.''}\footnote{
We assume a general inheritance axiom:\\
\m{isa(I,SuperC) :- isa(I,C), subclass\_of(C,SuperC).}} \\
\m{subclass\_of(computer,container).} \\ 
\ \\
{\it \% ``RAM size is a measure of capacity.''} \\
\m{capacity(X,Y) :-} \\
\m{~~~~~~~~isa(X,computer),} \\
\m{~~~~~~~~ram\_size(X,Y).} 
\end{my-quote}
However, this becomes problematic here as there is a second notion of
``computers as containers'' in our original axioms, namely computers
as containers of expansion cards.  If we map this notion onto our {\it
computer} theory in the same way, by adding the axiom:

\begin{my-quote}
{\it \% ``Number of expansion slots is a measure of capacity''} \\
\m{capacity(X,Y) :-} \\
\m{~~~~~~~~isa(X,computer),} \\
\m{~~~~~~~~expansion\_slots(X,Y).} 
\end{my-quote}
then the resulting representation captures that a computer has two
capacities (memory capacity and slot capacity), but loses the
constraints among their relations.  Consequently, memory capacity may
be used to compute the number of free expansion slots, and slot
capacity may be used to compute available RAM.  This illustrates how
the general container theory can be ``overlaid'' on a computer in
multiple ways, but inheritance fails to keep these overlays distinct.

This problem might be avoided in various ways.  We could insist that a
general theory (e.g., {\it container}) is applied at most once to a
more specific theory (although there is no obvious, principled justification
for this restriction).
 We would then revise our representation so that
it is not a computer, but a computer's {\it memory}, which contains
data, and similarly that a computer's {\it expansion slots} contain
cards.  While this solves the current problem, the general problem
remains.  For example, we may also want to model the computer's memory
as a container in other senses (e.g.,  of transistors, files,
information, or processes), which this restriction prohibits.

Another pseudo-solution is to parameterize the container theory,
by adding an argument to the {\it container}
axioms to denote the {\it type} of thing contained, to distinguish
different applications of the {\it container} theory.  With the
changes italicized, our axioms become:

\begin{my-quote}
{\it \% ``Free space for content-type T = capacity for T - occupied T.''} \\
\m{free\_space(Container,}{\it ContentType,}\m{F) :-} \\
\m{~~isa(Container,container),} \\
\m{~~capacity(Container,}{\it ContentType,}\m{C),} \\
\m{~~occupied\_space(Container,}{\it ContentType,}\m{O),} \\
\m{~~F is C - O.}\\
\ \\
{\it \% ``}\m{ram\_size} {\it denotes a computer's RAM capacity.''} \\
\m{capacity(X,{\it ram},Y) :-}  \\
\m{~~isa(X,computer),} \\
\m{~~ram\_size(X,Y).}
\end{my-quote}
Again, this solves the current problem (at the expense of parsimony),
but is not a good general solution.  Multiple parameters may be needed
to distinguish different applications of a general theory to a more
specific one.  For example, we would need to add a second parameter
about the container's \m{Dimension} (say) to distinguish physical
containment (as in: ``a computer contains megabytes of data'') from
metaphysical containment (as in: ``a computer contains valuable
information'').
This complicates our {\it container} axioms further, and still other
parameters may be needed.

A second limitation of inheritance is that it copies axioms (from a
general theory to a more specific one) in an ``all or nothing'' fashion.
Often only a selected part of a theory should be transferred.  To
continue with our example, the general {\it container} theory may
include relations for a container wall and its porosity, plus axioms
involving these relations.  Because the relations have no counterpart
in the {\it computer} theory, these relations and axioms should not be
transferred. 

These two problems arise because inheritance is being misused, not
because it is somehow ``buggy.''  When we say ``A computer is a
container,'' we mean ``A computer (or some aspect of it, such as its
memory) {\it can be modeled as} a container.''  Inheritance is
designed to transfer axioms through the {\it isa} relation, not the
{\it can-be-modeled-as} relation.  Nevertheless, knowledge engineers
often conflate these relations, probably because inheritance has been
the only approach available to them.  This leads to endless (and
needless) debates on the placement of abstract concepts in taxonomies. For
example, 
where should {\it container} be placed in a
taxonomy with respect to {\it object, substance, process} and so on?
Almost anything can be thought of as a container in 
some way, and if we pursue this route, we are drawn into debating these
modeling decisions as if they were issues of some objective 
reality.
This was a recurrent problem in our earlier work on the Botany
Knowledge-Base \cite{bkb}, where general theories used as models (such
as {\it connector} and {\it interface}) sit uncomfortably high in the
taxonomy.  The same issue arises in other ontologies.  For example,
{\it product} is placed just below {\it individual} in Cyc
\cite{cyc-ontology} and {\it place} is just below {\it
physical-object} in Mikrokosmos \cite{mikrokosmos}. 

\section{Knowledge Patterns \label{knowledge-patterns}}

Our approach for handling these situations is conceptually simple but
architecturally significant because it enables us to better modularize
a knowledge-base.  We define a {\it pattern} as a first-order theory
whose axioms are not part of the target knowledge-base, but can be
incorporated via a renaming of their non-logical symbols.

A theory acquires its status as a pattern by the way it is used,
rather than by having some intrinsic property.  First, the knowledge
engineer implements the pattern as an explicit, self-contained theory.
For example, the {\it container} theory would include the axiom:
\begin{my-quote} \begin{verbatim}
free_space(Container,F) :- 
    isa(Container,container),
    capacity(Container,C), 
    occupied_space(Container,O), 
    F is C - O.
\end{verbatim} \end{my-quote}
Second, using terminology from category theory \cite{categorytheory},
the knowledge engineer defines a {\it morphism} for each
intended application of this pattern in the target knowledge-base.
A morphism is a consistent\footnote{
Two examples of inconsistent mappings are: (i) mapping a symbol twice, e.g., 
\m{\{A->X,A->Y\}}, (ii) mapping a function \m{f} to \m{g}, where 
\m{g}'s signature as specified by
the mapping conflicts with \m{g}'s signature as already defined
in the target KB, e.g., \m{\{f->g,A->X,B->Y\}}, where 
$\m{f}:\m{A} \rightarrow \m{B}$ in the source pattern but $g$ is already
in the target and does not have signature $\m{g}:\m{X} \rightarrow \m{Y}$.}
mapping of the pattern's non-logical
symbols, or {\it signature}, to terms in the knowledge-base,
specifying how the pattern should be transformed. 
Finally, when the knowledge base is loaded, morphed copies of this pattern are
imported, one for each morphism. In our example, there are two
morphisms for this pattern:
\begin{my-quote} \begin{verbatim}
   container -> computer
   capacity -> ram_size
   free_space -> available_ram
   occupied_space -> occupied_ram
   isa -> isa
\end{verbatim} \end{my-quote}
\vspace{-0.5cm}
and 
\vspace{-0.3cm}
\begin{my-quote} \begin{verbatim}
   container -> computer
   capacity -> expansion_slots
   free_space -> free_slots
   occupied_space -> occupied_slots
   isa -> isa
\end{verbatim} \end{my-quote}
\vspace{-0.3cm}
(The reason for mapping a symbol to itself, e.g., the last line in
these morphisms, is explained in the next paragraph).
When these morphisms are applied, two copies of the {\it container}
pattern are created, corresponding to the two ways, described above,
in which computers are modeled as containers.

There may be symbols in the pattern that have no counterpart in the
target knowledge base, such as the thickness of a {\it container wall}
in our computer example.  In this event, the knowledge engineer omits
the symbols from the morphism, and the morphing procedure maps each
one to a new, unique symbol (generated by Lisp's gensym function, for
example).  This restricts the scope of these symbols to the morphed
copy of the pattern in the target knowledge base.  Although the
symbols are included in the imported theory, they are invisible (or
more precisely, hidden) from other axioms in the knowledge base.  Note
that we cannot simply delete axioms that mention these symbols because
other axioms in the imported theory may depend on
them.\footnote{Although specific axioms may be removed if they do not
contribute to assertions about symbols that are imported. A dependency
analysis algorithm could, in principle, identify and remove such
``dead code''.}  

\vspace{-0.5cm}

\begin{figure}

{\centering \centerline{\includegraphics[width=3.6in]{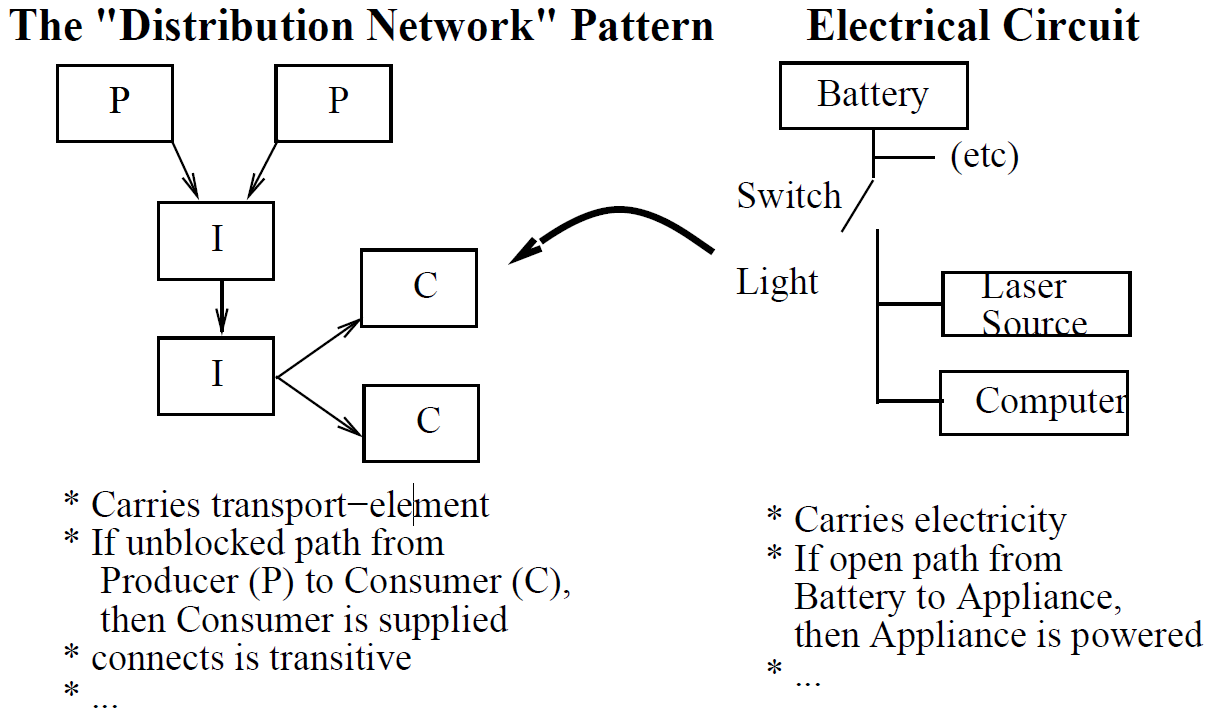}}}
\vspace{-0.4cm}
\caption{A knowledge pattern is created by abstracting the structure of
a theory (here, about electrical circuits). \label{distn-network}}
\end{figure}

\vspace{-0.4cm}

\begin{figure}
{\centering \centerline{\includegraphics[width=3.6in]{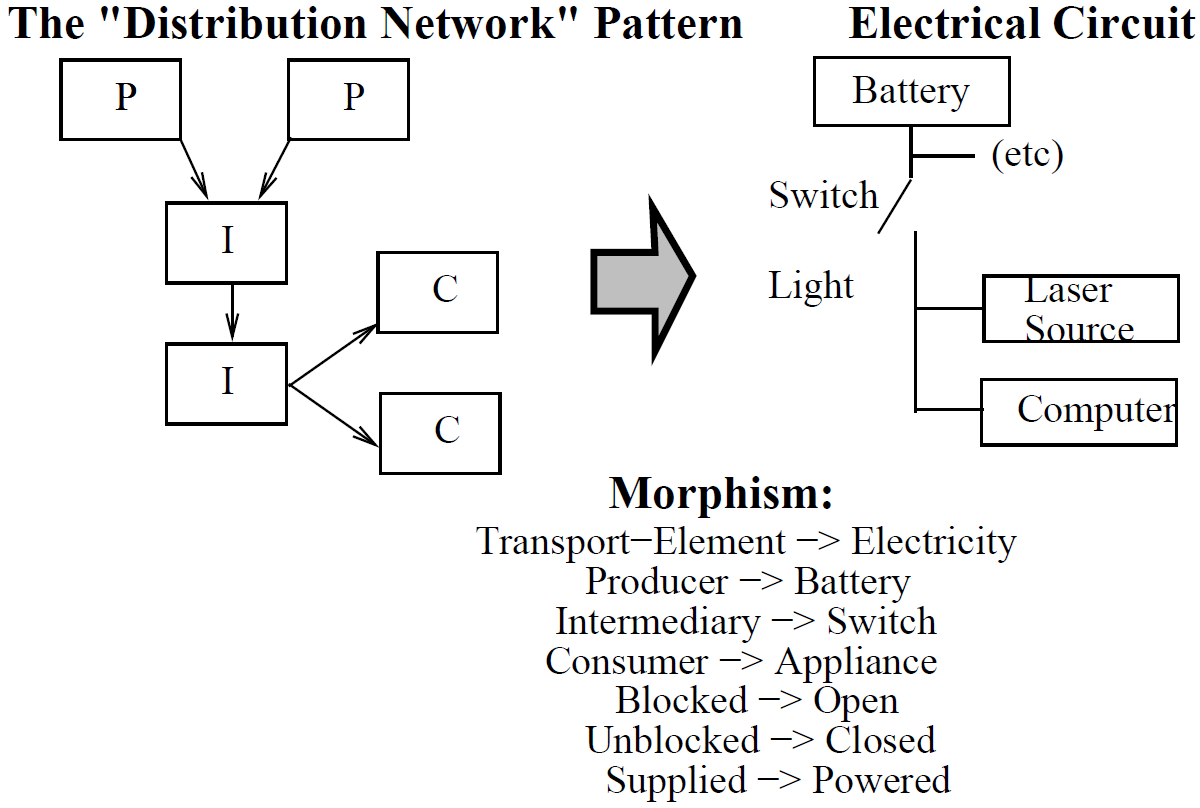}}}
\vspace{-0.3cm}
\caption{A knowledge pattern is applied by specifying
a mapping from symbols in the pattern to symbols in the target ontology
of interest. \label{distn-network2}}
\end{figure}


\input{fig1}		


\input{fig23}		


\section{Using Patterns for Building a Knowledge-Base \label{application}}

\noindent
We encountered the limitations of inheritance and developed the
approach of knowledge patterns while building KB-PHaSE, a prototype
knowledge-based system for training astronauts to perform a space
payload experiment called PHaSE (Physics of Hard Spheres Experiment).
PHaSE involves projecting a laser beam through various colloidal
suspensions of tiny spheres in liquids, to study the transitions among
solid, liquid, and glass (not gas) states in micro-gravity.
KB-PHaSE trains the astronaut in three ways.  First, it provides a
simple, interactive simulator in which the astronaut can step through
the normal procedure of the experiment.  Second, it introduces
simulated faults to train the astronaut to recover from problems.
Finally, it supports exploratory learning in which the astronaut can
browse concepts in the knowledge-base and ask questions using a
form-based interface. All three tasks use the underlying
knowledge-base to infer: properties of the current experimental state,
valid next actions, and answers to user's questions.
The prototype was built as a small demonstrator, rather than 
for in-service use, to provide input to Boeing and NASA's
Space Station Training Program. 
Details of KB-PHaSE are
presented in \cite{phase} and the
question-answering technology is described in \cite{qn-answering-symp}.

Our interest here is how the underlying knowledge-base was assembled
from component theories, rather than written from scratch.  KB-PHaSE
includes representations of many domain-specific objects (such as
electrical circuits) and processes (such as information flow) that are
derived from more general theories.  For example, we can think of an
electrical circuit in terms of a simple model of distribution, in
which producers (a battery) distribute a product (electricity) to
consumers (a light), illustrated schematically in 
Figures~\ref{distn-network} and~\ref{distn-network2}.
To capture this in a reusable way, we formulated
the general model of distribution as an independent, self-contained
pattern, shown in Figure~\ref{distribution-network}.
Then we defined a morphism that creates from it a model of electrical circuits, as
shown Figure~\ref{distribution-and-electrical-theories}. 

Our general theory of distribution was built, in turn, by extending a
general theory of blockable directed acyclic graphs (blockable-DAGs),
which in turn was built by extending a general theory of DAGs
(Figures~\ref{dag} and~\ref{blockable-dag}). 
The application, including these and 
other theories, 
is implemented in the frame-based language KM \cite{km-userman}.

\begin{figure}
{\centering \centerline{\includegraphics[width=3.3in]{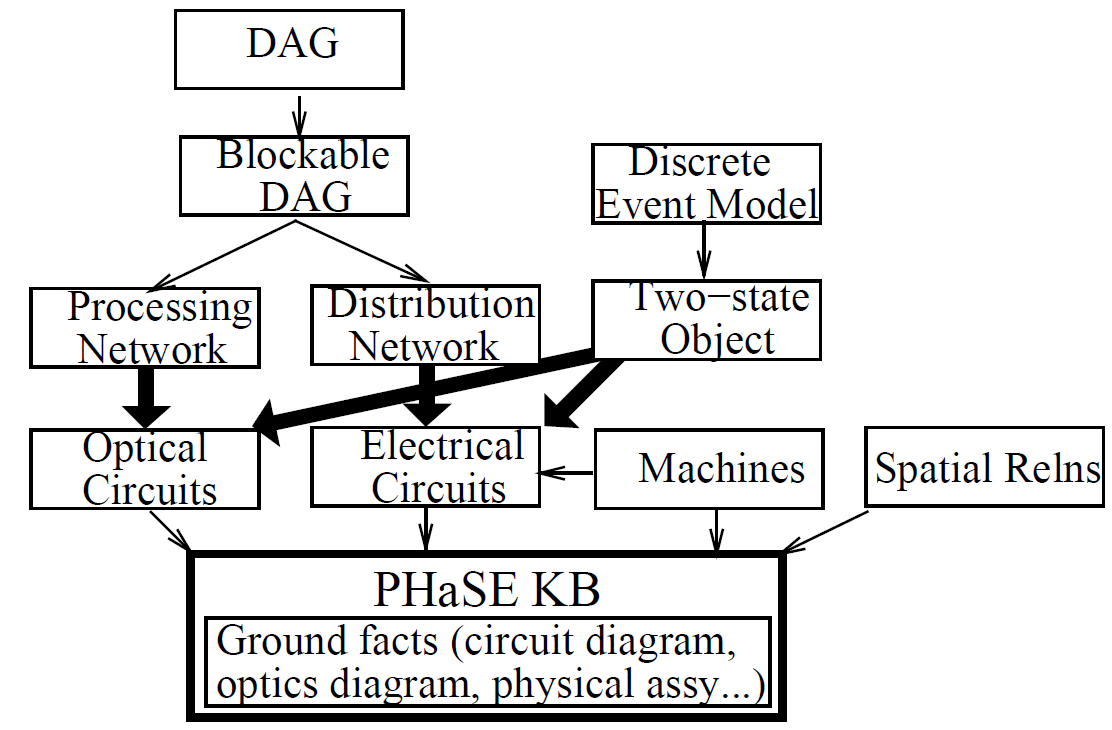}}}
\caption{The component theories used in KB-PHaSE. Each box denotes
        a theory (set of rules) describing a phenomenon, and arcs
	denote inclusion relations, the thick arcs involving morphing
	the source.
        \label{comp-arch}}
\end{figure}

By separating these theories as modular entities, they are available
for reuse.  In this application, we also modeled information flow in
the optical circuit (laser to camera to amplifier to disk) using a
morphed pattern describing a processing network, which, in turn, was
defined as an alternative extension of the basic blockable DAG theory,
thus reusing this theory. Similarly, the general pattern of a
``two-state object'' occurs several times within KB-PHaSE (e.g.,
switches, lights, and open/closed covers), and this pattern was again
made explicit and morphed into the knowledge base as required. These
patterns and their inter-relationships are shown in
Figure~\ref{comp-arch}.


\section{The Semantics of Knowledge Patterns}

A knowledge pattern is incorporated into a knowledge base by a syntactic 
process of symbol renaming (morphing). As the process is syntactic, 
it might seem difficult to provide semantics for the morphing process
itself. However, we can take some steps towards this by considering the
result of morphing to be logically equivalent to adding the knowledge pattern
directly into the knowledge base, along with some mapping axioms 
relating knowledge base terms to that knowledge pattern\footnote{We are
endebted to Richard Fikes for making this suggestion.}. If we can do this,
then those mapping axioms will have defined the semantics of what the morphing
operation has achieved. We provide here an outline of an approach to
doing this, although a full solution requires further work.

\begin{figure}[t]
{\centering \centerline{\includegraphics[width=4.55in]{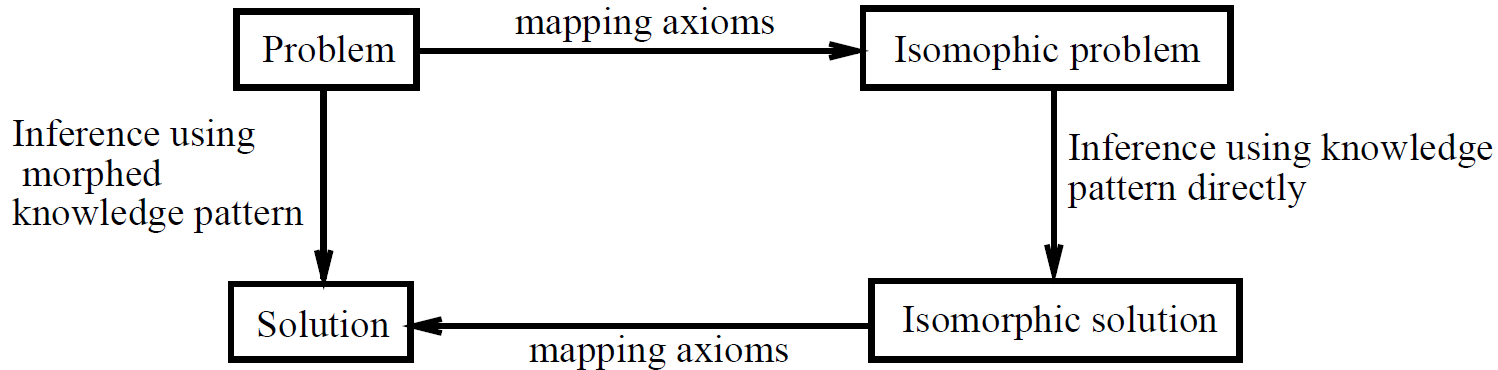}}}
\caption{To provide semantics for the syntactic process of morphing
(the left down arrow), 
we consider its equivalence to using the knowledge pattern 
directly, along with mapping axioms to translate a domain-specific problem 
into/out of an isomorphic one expressed using the pattern's ontology
(the other three arrows). \label{equivalence}}
\end{figure}

When morphing a knowledge pattern, we are ``bringing the abstract theory
to the application'', i.e., converting the vocabulary (ontology)
used in the pattern to that of the application domain in which it is 
to be used, via symbol renaming. 
An alternative, but functionally equivalent, approach would
be to take a domain-specific problem, and convert {\it its} vocabulary
to that used in the pattern, solve the problem using the pattern, and
then convert the result back to the domain-specific vocabulary. This
can be done, given a domain-specific problem, by establishing an isomorphic 
problem to solve using the pattern, solve it, and then translate the
results back.
This can be viewed as the complement to morphing, namely 
``bringing the application to the abstract theory.''
The approach is illustrated schematically in Figure~\ref{equivalence}.

This method is exactly that used in both object-oriented composition,
and reasoning by analogy. A classic example used in object-oriented 
composition (\cite{design-patterns}, pp18-22) is the task of
using a Rectangle concept to specify the area of a (graphics) Window.
Rather than doing this through inheritance, by stating that
a Window ISA Rectangle, the programmer states that a Window HAS a 
Rectangle instance associated with it. The Window object then delegates 
some queries (eg. its area) to the Rectangle, which computes the answer
and passes the answer back. In this example, the Window is the
application-specific object, while the Rectangle (along with its methods)
is equivalent to the knowledge pattern, and a domain-specific problem (e.g., 
the area of the Window) is solved by creating an isomorphic problem
(e.g., creating a Rectangle, and finding {\it its} area), and converting
the result back. A similar mechanism is used in reasoning by analogy,
where a problem is transferred between a base and a target theory
\cite{sme}.

If we can express this approach logically, we will have expressed the
semantics of a process equivalent to morphing, and hence provided a
semantics for morphing itself. In the earlier example of a computer, 
these ``mapping axioms'' would be:
\begin{quote}
{\it \% For each computer, assert that there exists an isomorphic container...} \\
$\forall x~ isa(x,Computer) \rightarrow \exists x'~ isa(x',Container) \wedge
		is$-$modeled$-$by_1(x,x')$ \\
\ \\
{\it \% ...whose spatial properties model the slot capacities of that computer.} \\
$\forall x,x'~ is$-$modeled$-$by_1(x,x') \rightarrow $ \\
$~~~~~~~~(~ \forall i~ capacity(x',i) \leftrightarrow expansion$-$slots(x,i) ~ \wedge$ \\
$~~~~~~~~~~ \forall j~ occupied$-$space(x',j) \leftrightarrow occupied$-$slots(x,j) ~ \wedge$ \\
$~~~~~~~~~~ \forall k~ free$-$space(x',k) \leftrightarrow free$-$slots(x,n) ~)$
\end{quote}
By adding the Container knowledge pattern directly into the KB,
along with these mapping axioms, the result will be the same as if we
had added a morphed version of the Container knowledge pattern:
\begin{center}
(~ Knowledge Pattern $\cup$ Mapping Axioms ~) $\equiv$ Morph(Knowledge Pattern)
\end{center}
Hence, in this case, the mapping axioms provide the semantics of what the 
morphing operation would have achieved. In the more general case, a 
domain-specific problem may involve multiple objects (e.g., an electrical 
circuit of electrical components), requiring setting up an isomorphic 
problem also involving multiple, corresponding objects expressed in the 
knowledge pattern's ontology. 

As this alternative approach is equivalent to morphing knowledge patterns,
why not simply use it, rather than morphing? This is a valid, alternative
approach to applying knowledge patterns, and achieves many of the same
goals (namely to make the abstract theories explicit, and to make explicit
the modeling decisions about how they apply to real-world phenomena).
The tradeoffs are largely computational, the mapping approach being 
more complex to implement and computationally more expensive at run-time,
but also having the advantage that the knowledge pattern itself is
then an explicit part of the final KB (rather than the KB containing only
morphed copies of that pattern).

\section{Related Work}

There are several important areas of pattern-related work, differing
in the type of reusable knowledge they encode and the way they encode
it.

In software engineering there has been considerable work on formal
methods for software specification, based on the construction and
composition of theories, and using category theory (applied to
algebraic specifications) as a mathematical basis (e.g.,
\cite{goguen,specware}).  SpecWare is an example of a software
development environment which is based on this approach and is capable
of synthesizing software semi-automatically \cite{specware-manual}.
As described in Section~\ref{knowledge-patterns}, our work can be
viewed as motivating, simplifying, and applying similar ideas to the
task of knowledge engineering.

Work on reusable problem-solving methods (PSMs), in particular KADS
\cite{kads} and generic tasks \cite{generictasks}, addresses
modularity and reuse in the context of procedural knowledge.  PSMs are
based on the observation that a task-specific method can be decomposed
into more primitive -- and more reusable -- sub-methods, and that
working with a library of such primitives may accelerate building a
system and make it more understandable and maintainable.  Work on PSMs
shares the same general goal that we have --- to identify and make
explicit recurring generalizations --- but it differs in two respects.
First, while PSMs are (mostly) patterns of procedural inference, we
have been targeting the basic domain knowledge (models) which those
procedures may operate on. (Although, since logic has both a
declarative and procedural interpretation, this distinction becomes
blurred).  Second, the mechanics of their usage differ:
implementations of PSMs can be thought of as parameterized procedures,
applied through instantiating their ``role'' parameters with domain
concepts (e.g., the ``hypotheses'' role in a diagnosis PSM applied to
medical diagnosis might be filled with disease types); in contrast,
our patterns are closer to schemata than procedures, and applied
instead through morphing.


Research on compositional methods for constructing ontologies and 
knowledge bases (eg \cite{ff,aaai97,noy-and-hafner}) has explored factoring
domain knowledge into component theories, analogous to factoring 
procedural knowledge into PSMs.
A component theory describes relationships among a set of 
objects (its participants) and is applied in an analogous
way to PSMs, by instantiating these participants with domain concepts.
Knowledge patterns develop this idea in two ways. First, they
provide further generalization, capturing the abstract structure 
of such theories. Second, their method of application differs 
(morphing, rather than axioms linking participants with 
domain concepts). This permits a pattern to be applied in multiple,
different ways to the same object, as discussed in 
Section~\ref{limits-of-inheritance}.
Compositional modeling has also explored the automated, run-time
selection of appropriate components to use \cite{ff,tripel}, 
an important issue which we have not addressed here.

``Design patterns'' in object-oriented programming 
(e.g., \cite{design-patterns}) are descriptions of common, useful
organizations of objects and classes, to help create specific
object-oriented designs. They again try to capture recurring
abstractions, but (in contrast to the approaches described earlier)
their primary intent is as architectural guidance to the software
designer, not as computational devices directly. As a result, they
are (and only need be) semi-formally specified, and they do not require
a method for their automatic application. (\cite{menzies} gives
an excellent discussion of the relationship between object-oriented
patterns and problem-solving methods). Another area of related
work from programming languages is the use of template programming
methods, where a code template is instantiated by syntactic
substitution of symbols within it (e.g., Ada generics, C++ templates),
corresponding to the syntactic implementation of pattern morphing,
but without the associated semantics.

In a similar way to design patterns, ``semantic patterns''
\cite{semantic-patterns1,semantic-patterns2} were intended as a means
of describing knowledge in an ``implementation neutral'' way, i.e., 
above the level of any particular representation language, with the
objective of reusing semantics across representational languages (in
particular, for Web-based applications). As with design patterns, they
are intended as a means for communication among human developers, but
in addition they contain various implementations of that knowledge,
expressed in different knowledge representation schemes. Semantic
patterns share some similarities with knowledge patterns, in particular
the goal of abstraction and reuse. However, they also are rather different
in other respects: semantic patterns were primarily intended for 
knowledge sharing across a wide range of representation systems, i.e.,
their language neutrality was a primary goal (with the consequence that
each implementation of the pattern has to be largely written by hand).
In contrast, knowledge patterns are intended for use within a particular
representational scheme in an automated way, with the primary goals of
uncovering and making explicit the abstract theories used in the knowledge 
base, and capturing the modeling decisions made about how these theories 
apply to the domain of interest.

Work on analogical reasoning is also closely related, as it similarly
seeks to use a theory (the base) to provide extra knowledge about 
some domain (the target), by establishing and using a mapping between the two. 
However, work on analogy has mainly focussed on identifying what the 
appropriate mappings between the base and target should be \cite{sme}, a
task which we have not addressed and which could be
beneficial for us to explore further. In addition, 
an alternative way of applying our patterns would
be to transform a domain-specific {\it problem} into the vocabulary
of a pattern (and solve it there, and transform the solution back), rather 
than transforming the pattern into the vocabulary of the domain.
In the PHaSE KB, for example, a query about the electrical circuit 
would be transformed to a query about a distribution network which
was isomorphic to the electrical circuit, solved there, and the
answer transformed back to the electrical circuit. This alternative
approach is similar to (one form of) solution by analogy, in which
the pattern (e.g., the distribution network) takes the role of 
the base, and the domain facts (e.g., the electrical circuit) the
target \cite{sme}. It is also similar to the use of delegation in 
object-oriented programming (the target `delegates' the problem to the
base, which solves it and passes the solution 
back \cite[p20]{design-patterns}). This variant
approach for using patterns would allow some run-time flexibility, 
but would be more complex to implement and computationally more
expensive at run-time.

Finally, work on  microtheories and contexts
(e.g., \cite{contexts-symposium,microtheories}) is also related, where
a microtheory (context) can be thought of as a pattern, and lifting axioms
provide the mapping between predicates in the microtheory and the
target KB which is to incorporate it. However, this work has typically
been used to solve a different problem, namely
breaking a large KB into a set of smaller, simpler (and thus more
maintainable) pieces, rather than making recurring axiom patterns explicit, and
it does not account for mapping the same microtheory multiple times
(and in different ways) into the same target KB. Reasoning with
lifting axioms can also be computationally expensive except in the
simplest cases. 

Note that patterns are not an essential prerequisite for 
building a knowledge-based system. In the PHaSE application,
for example, we could have simply defined the PHaSE electrical 
circuit, implemented axioms about the behavior of electrical
circuits, and answered circuit questions, all within the electrical
vocabulary. This would be a completely reasonable approach for
a single-task system; however, to achieve reuse within a 
multifunctional system (such as KB-PHaSE), or between systems,
it becomes preferable to extract the more general abstractions,
as we have described. Patterns do not 
enable better reasoning, rather they are to help reuse.

\section{Summary}

Ontological engineering is fundamentally a modeling endeavor. In this
Chapter, we have described a knowledge engineering technique aimed at 
helping in this endeavor, by making recurring theory schemata, or knowledge
patterns, explicit, and available for manipulation. From a computational
perspective, knowledge patterns provide a simple and computationally
efficient mechanism for facilitating knowledge reuse. From a modeling 
perspective, knowledge patterns also provide an important insight into
the process of ontological engineering, namely that it is not simply
about ``writing axioms'', but also involves recognizing that 
the ``computational clockwork'' of one or more abstract theories seem to 
behave (to a reasonable approximation) in the same way as some
system of objects in the world, and hence can be used to describe it.
Knowledge patterns make both those abstract theories and their mappings
to the domain of interest explicit, thus making modeling decisions clear,
and avoiding some of the ontological confusion that can otherwise arise.

However, our approach also has limits. First, it does not allow a system to make
run-time modeling decisions, as general theories are morphed when the
knowledge base is loaded. Second, it does not address the
issue of {\it finding} relevant knowledge patterns in the first
place, or deciding the appropriate boundaries of patterns
(this is left to the knowledge engineer). Finally, we do not 
address the issue of finding the appropriate mappings between
patterns and the domain; this again is left to the knowledge
engineer. As mentioned earlier, this is a primary focus of research in 
the related field of analogical reasoning \cite{sme}.

Despite these,
the significance of this approach is that it allows us to better 
modularize the axioms which underly formal ontologies, 
and isolate general theories as self-contained units for reuse. 
It also allows us to control and vary the way those theories are mapped 
onto an application
domain, and it better separates the ``computational clockwork'' of a 
general theory from the domain phenomena which it is considered to 
reflect. In addition, the approach is technically simple and
not wedded to a particular implementation language.
In the long-term, we hope this will help foster 
the construction of reusable theory libraries, an essential
requirement for the construction of large-scale, formal ontologies
and  knowledge-based systems.





\input{handbook03.bbl}


\end{document}

%% file: mydefs.tex
\newenvironment{mylist}{                        
        \parskip 0cm \begin{list}{}{\parsep 0cm \itemsep 0cm \topsep 0cm}}{
        \end{list} \parskip 0cm}
\newcommand{\indexentry}[2]{#1 #2 \\}
\newcommand{\bt}{~\mbox{${\bullet}$~}}          
\newcommand{\ra}{\mbox{$\rightarrow$}}
\newcommand{\rt}{\mbox{$\rightarrow$~}}
\newcommand{\tl}[1]{\begin{center} {\Large \bf #1} \end{center}}

\newcommand{\comment}[1]{{\tiny{\it Comment: #1}}}

\newcommand{\fa}{$\forall$}
\newcommand{\ex}{$\exists$}
\newcommand{\im}{$\rightarrow$}
\newcommand{\an}{$\wedge$}
\newcommand{\ur}{$\vee$}
\newcommand{\da}{\mbox{-}}
\newcommand{\co}{\!:\!}

\newcommand{\m}[1]{{\mbox{\tt #1}}}           
\newcommand{\x}[1]{\mbox{\tt #1}}	
\newcommand{\f}[1]{{\bf #1}\index{\nil{#1}{\bf #1}}}	
\newcommand{\qf}[1]{\index{\nil{#1}{\bf #1}}}		
\newcommand{\p}[1]{{\tt #1}\index{\nil{#1}{\tt #1}}}  	
\newcommand{\qp}[1]{\index{\nil{#1}{\tt #1}}}  		
\newcommand{\ix}[1]{#1\index{\nil{#1}#1}}  		
\newcommand{\qi}[1]{\index{\nil{#1}#1}}  		
\newcommand{\qix}[1]{\index{\nil{#1}#1}}  		
\newcommand{\sm}[1]{\mbox{\small ${\tt #1}$}}           
\newcommand{\sd}[1]{\footnotesize \mbox{$\sigma #1$}}
\newcommand{\se}[1]{\footnotesize \mbox{$\pm #1$}}
\newcommand{\nil}[1]{}
\newenvironment{mydescription}{                 
     \parskip 0cm \begin{description}{\parsep 0cm \itemsep 0cm \topsep 0cm}}{
        \end{description}} 
\newenvironment{myquote}{                 
     \begin{quote} \small \parskip 0cm \topsep 0cm}
        {\end{quote} \normalsize} 
\newenvironment{myenumerate}{                   
     \parskip 0cm \begin{enumerate}{\parsep 0cm \itemsep 0cm \topsep 0cm}}{
        \end{enumerate}} 
\newenvironment{myitemize}{                     
     \parskip 0cm \begin{itemize}{\parsep 0cm \itemsep 0cm \topsep 0cm}}{
        \end{itemize}} 
\newenvironment{des}{                 
     \parskip 0cm \begin{list}{}{\parsep 0cm \itemsep 0cm \topsep 0cm}}{
       \end{list}} 
\newenvironment{enu}{                   
     \parskip 0cm \begin{list}{}{\parsep 0cm \itemsep 0cm \topsep 0cm}}{
       \end{list}} 
\newenvironment{ite}{                     
     \parskip 0cm \begin{itemize}{\parsep 0cm \itemsep 0cm \topsep 0cm}}{
        \end{itemize}} 
\newenvironment{ver}{\parskip 0cm \small \begin{verbatim}}{\end{verbatim} \normalsize}
\newcommand{\ti}[1]{\begin{center} {\bf #1} \end{center}}
\newcommand{\s}[1]{\section{#1}}
\renewcommand{\ss}[1]{\subsection{#1}}
\newcommand{\sss}[1]{\subsubsection{#1}}
\newcommand{\ssss}[1]{\noindent {\bf #1}}
\newcommand{\myhndrawscale}[2]{\vspace{1cm} \hndrawscale{#1}{#2}}
\newcommand{\myyhndrawscale}[2]{\vspace{0.5cm} \hndrawscale{#1}{#2}}
\newcommand{\myline}{\vspace{-0.4cm}\begin{center} \underline{~~~~~~~~~~~~~~~~~~~~~~~~~~~~~~~~~~~~~~~~} \end{center}\vspace{-0.6cm}}

%% file: fig1.tex

\begin{figure*}[t]
\fbox{
\begin{minipage}{4.55in}
\begin{center}
{\large \bf Pattern: DAG (Directed Acyclic Graph)}
\end{center}
{\centering \centerline {\includegraphics[width=2in]{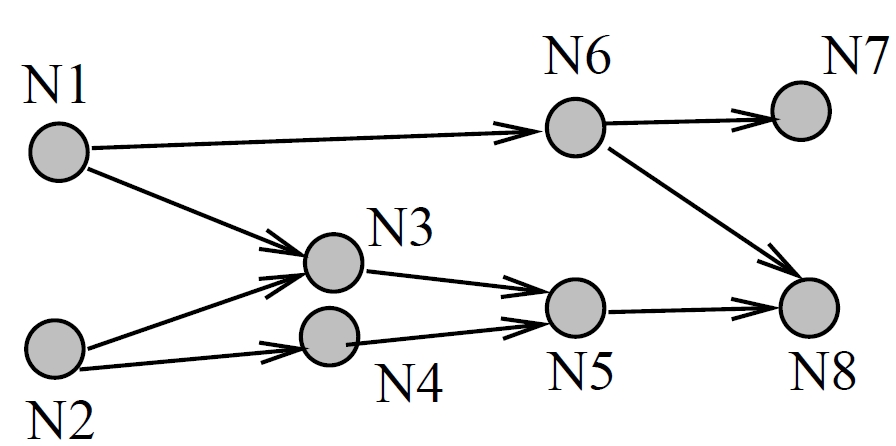}}}
\vspace{-1.2cm}
\subsection*{Synopsis}
\vspace{-0.3cm}
\begin{tabular}{p{3.5in}p{2in}}
{\bf Name:} \m{dag} \\
{\bf Summary:} Core theory of directed acyclic graphs. \\
{\bf Uses:} (none) \\
{\bf Used by:} \m{blockable-dag} \\
& 
\end{tabular}

\vspace{-0.3cm}

\noindent
{\bf Description:}
This component provides a basic axiomatization of DAGs, a
fundamental structure for modeling many real-world phenomena.
In a DAG, a NODE is directly linked TO and FROM zero or more other
nodes [1]. 
A node REACHES all its downstream nodes [2] \& is 	
REACHABLE-FROM all its upstream nodes [3].


\noindent
{\bf Signature:} $Node$, $DAG$, $node$-$in$, $to$, $from$, $reaches$, $reachable$-$from$, $isa$.

\vspace{-0.6cm}

\subsection*{Axioms:}
\vspace{-0.1cm}

\begin{tabular}{ll}
$\forall x,y~ to(x,y) \rightarrow isa(x,Node) \wedge isa(y,Node)$ & [1] \\
$\forall x,y~to(x,y) \leftrightarrow from(y,x)$\\
$\forall x,y~ to(x,y) \rightarrow reaches(x,y)$ & [2] \\
$\forall x,y,z~ to(x,y) \wedge reaches(y,z) \rightarrow reaches(x,z)$ \\
$\forall x,y~ from(x,y) \rightarrow reachable$-$from(x,y)$ & [3]\\
$\forall x,y,z~from(x,y) \wedge reachable$-$from(y,z) \rightarrow reachable$-$from(x,z)$  \\
$\forall x,y~isa(x,DAG) \wedge node$-$in(y,x) \rightarrow isa(y,Node)$  
\end{tabular}
\end{minipage}
}			
\caption{A knowledge pattern used in KB-PHaSE. \label{dag}}
\end{figure*}




\begin{figure*}[t]

\fbox{
\begin{minipage}{4.55in}
\begin{center}
{\large \bf Pattern: Blockable-DAG}
\end{center}
{\centering \centerline {\includegraphics[width=2in]{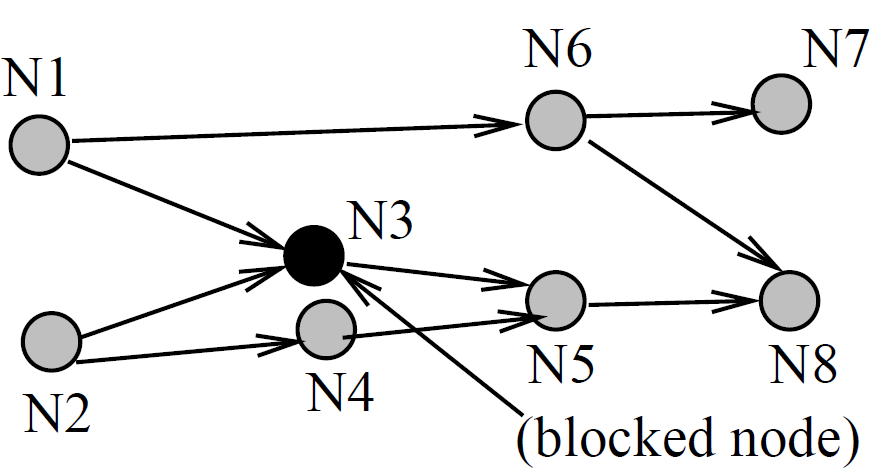}}}
\vspace{-0.8cm}
\subsection*{Synopsis}
\vspace{-0.2cm}
\begin{tabular}{p{3.5in}p{2in}}
{\bf Name:} \m{blockable-dag} \\
{\bf Summary:} Extension to DAG theory, in which nodes can be blocked (preventing reachability). \\
{\bf Uses:} \m{dag} \\
{\bf Used by:} \m{distribution-network} \\
& 
\end{tabular}

\vspace{-0.3cm}

\noindent
{\bf Description:}
A NODE may be BLOCKED or UNBLOCKED [1]. A node UNBLOCKED-REACHES
a downstream node if there is a path of UNBLOCKED 
nodes connecting the two [2].


\noindent
{\bf Signature:} That for \m{dag}, plus 
$blocked$, $unblocked$,
$unblocked$-$directly$-$reaches$, $unblocked$-$directly$-$reachable$-$from$,
$unblocked$-$reaches$, $unblocked$-$reachable$-$from$,


\noindent
{\bf Axioms:} \m{dag} theory axioms, plus: \\
\begin{tabular}{ll}
$\forall x~ isa(x,Node) \rightarrow blocked(x) \vee unblocked(x)$ & \hspace{-2cm} [1] \\
$\forall x~ blocked(x) \leftrightarrow \neg unblocked(x)$ \\
$\forall x,y~ to(x,y) \wedge \neg blocked(y) \rightarrow unblocked$-$directly$-$reaches(x,y)$ \\
$\forall x,y~ unblocked$-$directly$-$reaches(x,y) \rightarrow unblocked$-$reaches(x,y)$ & \hspace{-2cm} [2] \\
$\forall x,y,z~unblocked$-$directly$-$reaches(x,y) ~\wedge$ \\
$~~~~~~~~~~unblocked$-$reaches(y,z) \rightarrow unblocked$-$reaches(x,z)$ \\
$\forall x,y~ from(x,y) \wedge \neg blocked(y) \rightarrow unblocked$-$directly$-$reachable$-$from(x,y)$ \\
$\forall x,y~ unblocked$-$directly$-$reachable$-$from(x,y) \rightarrow unblocked$-$reachable$-$from(x,y)$ \\
$\forall x,y,z~ unblocked$-$directly$-$reachable$-$from(x,y) ~\wedge$ \\
$~~~~~unblocked$-$reachable$-$from(y,z) \rightarrow unblocked$-$reachable$-$from(x,z)$ \\
\end{tabular}
\end{minipage}
}			
\caption{Another knowledge pattern used in KB-PHaSE. \label{blockable-dag}}
\end{figure*}


%% file: fig23.tex


\begin{figure*}[t]
\fbox{
\begin{minipage}{4.55in}
\begin{center}
{\large \bf Pattern: Distribution Network}
\end{center}
{\centering \centerline {\includegraphics[width=2in]{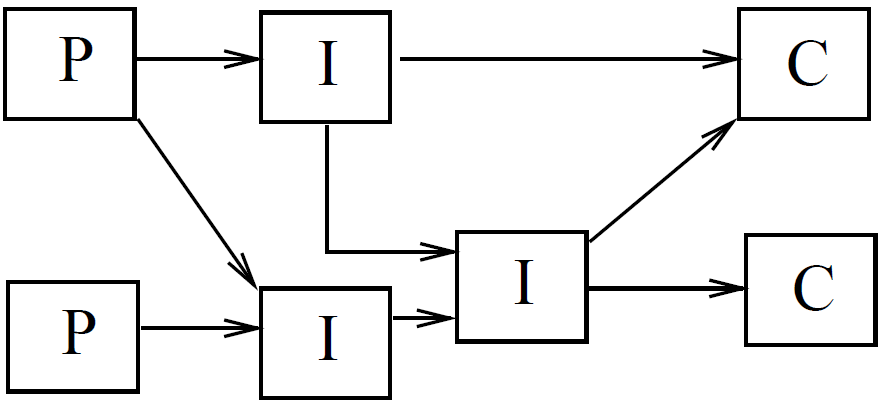}}}
\vspace{-1cm}
\subsection*{Synopsis}
\vspace{-0.2cm}
\begin{tabular}{p{3.5in}p{2in}}
{\bf Name:} \m{distribution-network} \\
{\bf Summary:} Simple theory of producers, intermediaries and
	consumers. \\
{\bf Uses:} \m{blockable-dag} \\
{\bf Used by:} \m{electrical-circuit} \\
& 
\end{tabular}

\vspace{-0.4cm}



{\bf Description:} 
A distribution network consists of three classes 
of nodes: PRODUCER, CONSUMER, and INTERMEDIARY [1], and the type of item 
transported is denoted by TRANSPORT-MATERIAL-TYPE
(e.g., \m{Water}) Examples include: electrical circuits, hydraulic 
circuits, commuter traffic.

In this model, there is a flow of TRANSPORT-MATERIAL-TYPE from PRODUCERs
to CONSUMERs via INTERMEDIARYs, providing the intermediary is not BLOCKED.
A CONSUMER/INTERMEDIARY is SUPPLIED if there is at least one UNBLOCKED 
path to it from a SUPPLIER [2]. All elements in the network transport that
network's TRANSPORT-MATERIAL-TYPE [3].


%


\noindent
{\bf Signature:} That for \m{blockable-dag}, plus 
$Producer$, $Consumer$, $Intermediary$, $Transport$-$Material$-$Type$, 
$supplied$, $product$-$type$, $consumes$-$type$.


\noindent
{\bf Axioms:} \m{blockable-dag} theory axioms, plus: \\
\begin{tabular}{ll}
$\forall x~isa(x,Producer) \rightarrow isa(x,Node)$ & [1] \\
$\forall x~isa(x,Consumer) \rightarrow isa(x,Node)$ \\
$\forall x~isa(x,Intermediary) \rightarrow isa(x,Node)$ \\
$\forall x~isa(x,Consumer) \wedge
(~\exists y~isa(y,Producer) \wedge$ \\
$~~~~~~~~~unblocked$-$reaches(y,x)~) \rightarrow supplied(x)$ & [2] \\
$\forall x~isa(x,Producer) \rightarrow product$-$type(x,Transport$-$Material$-$Type)$ & [3] \\
$\forall x~isa(x,Consumer) \rightarrow consumes$-$type(x,Transport$-$Material$-$Type)$ \\
\end{tabular}

\end{minipage}
}			
\vspace{0.2cm}

\caption{The knowledge pattern for distribution networks, used by KB-PHaSE. \label{distribution-network}}
\end{figure*}

\begin{figure*}[t]
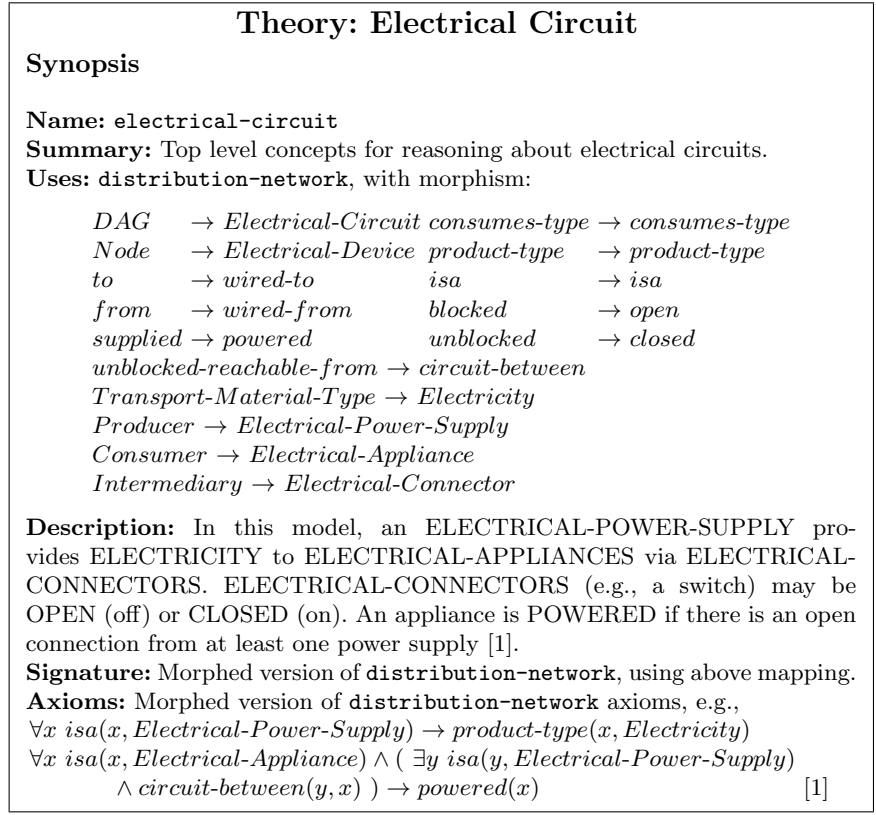

\fbox{
\begin{minipage}{4.35in}
\begin{center}
{\large \bf Theory: Electrical Circuit}
\vspace{-0.5cm}
\end{center}
\subsection*{Synopsis}
\noindent
{\bf Name:} \m{electrical-circuit} \\
{\bf Summary:} Top level concepts for reasoning about electrical circuits. \\
{\bf Uses:} \m{distribution-network}, with morphism:
\begin{center}
\begin{tabular}{llllll}
     $DAG$ & $\rightarrow$ & $Electrical$-$Circuit$ & 	$consumes$-$type$ & $\rightarrow$ & $consumes$-$type$ \\
     $Node$ & $\rightarrow$ & $Electrical$-$Device$ &   	$product$-$type$ & $\rightarrow$ & $product$-$type$ \\
     $to$ & $\rightarrow$ & $wired$-$to$ &      $isa$ & $\rightarrow$ & $isa$ \\
     $from$ & $\rightarrow$ & $wired$-$from$ &       $blocked$ & $\rightarrow$ & $open$  \\
     $supplied$ & $\rightarrow$ & $powered$ &    $unblocked$ & $\rightarrow$ & $closed$ \\
\multicolumn{6}{l}{$unblocked$-$reachable$-$from$   $\rightarrow$   $circuit$-$between$} \\
\multicolumn{6}{l}{$Transport$-$Material$-$Type$   $\rightarrow$   $Electricity$} \\
\multicolumn{6}{l}{$Producer$   $\rightarrow$   $Electrical$-$Power$-$Supply$} \\
\multicolumn{6}{l}{$Consumer$   $\rightarrow$   $Electrical$-$Appliance$} \\
\multicolumn{6}{l}{$Intermediary$   $\rightarrow$   $Electrical$-$Connector$}
\end{tabular}
\end{center}


{\bf Description:} In this model, an ELECTRICAL-POWER-SUPPLY provides 
ELECTRICITY to ELECTRICAL-APPLIANCES via ELECTRICAL-CONNECTORS.
ELECTRICAL-CONNECTORS (e.g., a switch) may be OPEN (off) or 
CLOSED (on). An appliance is POWERED if there is an open
connection from at least one power supply [1].


\noindent
{\bf Signature:} Morphed version of \m{distribution-network}, using above mapping.


\noindent
{\bf Axioms:} Morphed version of \m{distribution-network} axioms,  e.g., \\
\begin{tabular}{ll}
$\forall x~isa(x,Electrical$-$Power$-$Supply) \rightarrow product$-$type(x,Electricity)$ \\
$\forall x~isa(x,Electrical$-$Appliance) \wedge (~\exists y~isa(y,Electrical$-$Power$-$Supply)$ \\
$~~~~~~~~~~\wedge circuit$-$between(y,x)~) \rightarrow powered(x)$ & [1] \\
\end{tabular}
\end{minipage}
}			

\caption{The theory for electrical circuits in KB-PHaSE,
defined as a morphism of the ``distribution network'' knowledge pattern.
\label{distribution-and-electrical-theories}}
\end{figure*}



%% file: handbook03.bbl
\begin{thebibliography}{99.}

\bibitem{phase}
Clark P, Thompson J, Dittmar M (1998)
\newblock {KB-PHaSE}: A knowledge-based training tool for a space station
  experiment.
\newblock Technical Report {SSGTECH}-98-035, Boeing Applied Research and
  Technology, Seattle, {WA}

\bibitem{bkb}
Porter B, Lester J, Murray K, Pittman K, Souther A, Acker L, Jones T (1988)
\newblock {AI} research in the context of a multifunctional knowledge base: The
  botany knowledge base project.
\newblock Tech Report AI-88-88, Dept CS, Univ Texas at Austin

\bibitem{cyc-ontology}
{Cycorp, Inc.} (1996)
\newblock The cyc public ontology.
\newblock (http://www.cyc.com/public.html)

\bibitem{mikrokosmos}
Mahesh K, Nirenberg S (1995)
\newblock A situated ontology for practical {NLP}.
\newblock In:Proc. IJCAI-95 Workshop on Basic Ontological Issues in
  Knowledge Sharing

\bibitem{categorytheory}
Pierce B (1991) 
\newblock Basic Category Theory for Computer Scientists.
\newblock {MIT Press}

\bibitem{qn-answering-symp}
Clark P, Thompson J, Porter B (1999)
\newblock A knowledge-based approach to question-answering.
\newblock In: Fikes R, Chaudhri V (eds) Proc. {AAAI}'99 Fall
  Symposium on Question-Answering Systems. {AAAI}

\bibitem{km-userman}
Clark P, Porter B (1999).
\newblock {KM} -- the knowledge machine: Users manual.
\newblock Technical report, {AI} Lab, Univ Texas at Austin

\bibitem{design-patterns}
Gamma E, Helm R, Johnson R, Vlissides J (1995)
\newblock Design Patterns.
\newblock Addison-Wesley

\bibitem{sme}
Falkenhainer B, Forbus K, Gentner D (1986)
\newblock The structure-mapping engine.
\newblock In: AAAI-86, pages 272--277

\bibitem{goguen}
Goguen J (1986)
\newblock Reusing and interconnecting software components.
\newblock In: Computer 19(2):16--28

\bibitem{specware}
Srinivas Y V, Jullig, R (1995)
\newblock Specware: Formal support for composing software.
\newblock In: Proc. Conf. on the Mathematics of Program Construction,
  Kloster Irsee, Germany
\newblock (Also Kestrel Tech Rept {KES.U.94.5},
  http://www.kestrel.edu/HTML/publications.html)

\bibitem{specware-manual}
Jullig R, Srinivas Y, Blain L, Gilham L, Goldberg A, Green C, McDonald J, Waldinger R (1995)
\newblock Specware language manual.
\newblock Technical report, Kestrel Institute

\bibitem{kads}
Wielinga B J, Schreiber A T, Breuker J A (1992)
\newblock {KADS}: A modelling approach to knowledge engineering.
\newblock Knowledge Acquisition 4(1)

\bibitem{generictasks}
Chandrasekaren B (1986)
\newblock Generic tasks in knowledge-based reasoning: High-level building
  blocks for expert system design.
\newblock IEEE Expert, pages 23--30

\bibitem{ff}
Falkenhainer B, Forbus K (1991)
\newblock Compositional modelling: Finding the right model for the job.
\newblock Artificial Intelligence 51:95--143

\bibitem{aaai97}
Clark P, Porter B (1997)
\newblock Building concept representations from reusable components.
\newblock In: {AAAI}-97, pages 369--376, {CA:AAAI}

\bibitem{noy-and-hafner}
Noy N, Hafner, C (1998)
\newblock Representing scientific experiments: Implications for ontology design
  and knowledge sharing.
\newblock In: {AAAI}-98, pages 615--622

\bibitem{tripel}
Rickel J, Porter B (1997)
\newblock Automated modeling of complex systems to answer prediction questions.
\newblock Artificial Intelligence 93(1-2):201--260

\bibitem{menzies}
Menzies T (1997)
\newblock Object-oriented patterns: Lessons from expert systems.
\newblock Software -- Practice and Experience 27(12):1457--1478

\bibitem{semantic-patterns1}
Staab S, Erdmann M, Maedche A (2001)
\newblock Engineering ontologies using semantic patterns.
\newblock In: Preece A, {O'Leary} D (eds) Proc. {IJCAI-01}
  Workshop on e-Business and the Intelligent Web

\bibitem{semantic-patterns2}
Staab S, Erdmann M, Maedche A (2001)
\newblock Semantic patterns.
\newblock Technical report, Univ. Karlsruhe

\bibitem{contexts-symposium}
Buvac S (ed) (1995)
\newblock Proc AAAI-95 Fall Symposium on Formalizing Context. CA:AAAI

\bibitem{microtheories}
Blair P, Guha R, Pratt W (1992)
\newblock Microtheories: An ontological engineer's guide.
\newblock Tech Rept CYC-050-92, MCC, Austin, TX

\end{thebibliography}
